# An Early-Stage Workflow Proposal for the Generation of Safe and Dependable AI Classifiers

Hans Dermot Doran, Suzana Veljanovska
*Institute of Embedded Systems*
*Zurich University of Applied Sciences*
Winterthur, Switzerland
{donn, veln} @zhaw.ch

*Abstract*—The generation and execution of qualifiable safe and dependable AI models, necessitates definition of a transparent, complete yet adaptable and preferably lightweight workflow. Given the rapidly progressing domain of AI research and the relative immaturity of the safe-AI domain the process stability upon which functionally safety developments rest must be married with some degree of adaptability. This early-stage work proposes such a workflow basing it on a an extended ONNX model description. A use case provides one foundations of this body of work which we expect to be extended by other, third party use-cases.

*Keywords—safe-AI, dependable-AI, functional safety, AI workflows*

I. INTRODUCTION

The generation and execution of qualifiable safe, dependable or reliable AI-classifiers, our particular interest is in deep neural networks, necessitates a universally agreed upon workflow. We maintain that the functional safety domain represents the state of the art in workflow stringency and so to our understanding the acceptance of such workflows is dependent on coherence with established patterns from the functional safety domain [1] which themselves are derived from best practices from the more general engineering domain.

Developing functional safety (FS) systems involves achieving extremely low probabilities of failures for both hardware and software. Typical applications follow the sample-compute-actuate paradigm. In FS, dependable or reliable systems, the parameters which together with sampled values are used to calculate actuator values by the compute algorithms have meaning, are understood and are justifiable to certification and qualification authorities.

The AI-workflow differs significantly in that the parameters of the algorithm are derived from a training system, often using data from indeterminate sources. The semantic origin of these parameters is not well understood and hence not justifiable to an independent authority. This fact is the basis of the research domain of explainable AI [2].

For highly complex and context-rich domains, automated driving for instance, there is a growing trend to qualify the output of AI classifiers by other AI means, for instance classical expert systems [3]. For many other applications less dependent on context we maintain that the number of classes that must be determined reliably is small with respect to the total number of classes an AI-classifier can potentially identify. We have proposed a solution for such applications and in order to systematically progress this proposal we require a suitable AI workflow. We briefly illustrate our approach to generating reliable AI classifiers and based on that we propose, in Section II, a suitable workflow for further discussion. We describe early work and draw appropriate conclusions in Section III.

*A. Motivation*

In previous work we have postulated on the use of the single protected channel pattern (SPCP, [4]) for reliable validation of CNN outputs [5]. An exhaustive analysis is still outstanding, not least because we lack a suitable workflow to structure this analysis. It is this lack of workflow that motivates us to commence discussion of features of an AI workflow for functional safety/dependability or reliability.

We are unaware of any publications discussing workflows for generating and executing safe and dependable deep neural networks.

*B. Motivating Use-Case Description*

In safe and dependable systems due consideration has been given to methods of ingress of non-reliable data into a reliable system. The single protected channel pattern is one method by which to validate data provided by a non-reliable unit, be it a sensor or data from a similarly non-reliable source. We apply, and modify, this general pattern to achieve reliable AI execution as shown in Figure 1.

Our use-case is an industrial one where we are required to validate objects by simple shape recognition. It is convenient for us to use AlexNet as a convolutional neural network. We train the CNN with images of various objects and for shape validation we use Symbolic Aggregate Approximation (SAX, [6]). Since the input to the SAX algorithm is a binarized shape image the first step in preparing the data is edge recognition using a Sobel filter. Conveniently we can utilise one kernel of the 96 stipulated by AlexNet to produce the edges. By executing this reliably we can feed the output of this stage into both the validator and the CNN (c1f1 in Figure 1 below). We refer the reader for further details to [5]. For the purposes of this body of work this demonstrates that the smallest task that might be executed reliably is, at minimum a convolution kernel.

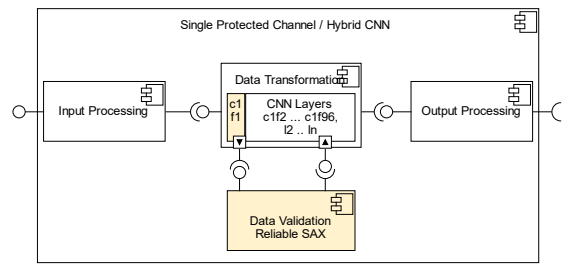

*Figure 1: Single Protected Channel Pattern Applied to AI Classifiers*

We thank SERI, Innosuisse and the ZHAW for the financial support of this research under the umbrella of the KDT project REBECCA, grant agreement n° 101097224.

With this information we are in a position to define an alpha version of a workflow on which we elaborate in the following section.

## II. Workflow Proposal

Having established the baseline criteria for an execution we progress to our workflow proposal. The fundamental philosophy behind this proposal is that the workflow should be qualifiable which in turn implies the use of qualifiable tools. The act of qualifying tools engenders considerable effort and expense. It follows that any updates beyond a certain complexity will require (expensive) re-qualification. We therefore place value on a lightweight workflow featuring as few as possible tools with limited features.

The application of our articulation of the single protected channel pattern puts the focus of dependability on the validator rather than the CNN. This in turn means tools such as PyTorch or TensorFlow can be used to train models but need not be safety qualified. We nevertheless require a model representation that is independent of one of these tools. We choose the ONNX format noting that it is supported by a relatively neutral organisation and it appears that extensions to what is a de facto standard can be easily integrated.

Based on this decision we refer to Figure 2, yellow/shaded regions refer to qualifiable tools. We propose definition of the model in ONNX format (Figure 2, .onnx file A). This file can be used to train the algorithm and produce an output ONNX representation (Figure 2, .onnx file B). We first validate that both files A and B have the same architecture, the qualified validator produces a qualified ONNX representation encoded in (Figure 2, .onnx file C). This file and an input of reliability attributes serves as input to the Model Partitioner which generates two files – one for runtime execution on a reliable platform (Figure 2, .onnx file D) and one for runtime execution on non-reliable platforms (Figure 2, .onnx file E). The partitioner will also accept an attribute file with directions as to how the model is to be partitioned. The Architecture Validator must be able to validate that the two runtime files when recombined produce the intended CNN architecture. It is still under discussion whether the validator (of the SPCP) should be included as part of the ONNX representation or not.

## III. Conclusions and Further Work

### A. Impact

As we are unaware of any relevant prior work we consider our proposal, a workflow that facilitates the design, generation and execution a novelty and our contribution is to initiate extending the state of the art in this domain.

### B. Further Work – Output

We are currently finalising the design for a prototype of the two tools (Architecture Validator and Partitioner) and intend to implement it in a programming language that facilitates code inspection and provability. We believe Scheme to be a suitable candidate.

### C. Further Work – Outcome

When the prototypes are completed and the workflow proven for our use-case, it is our intention to suggest reliability and traceability enhancements to the ONNX standard.

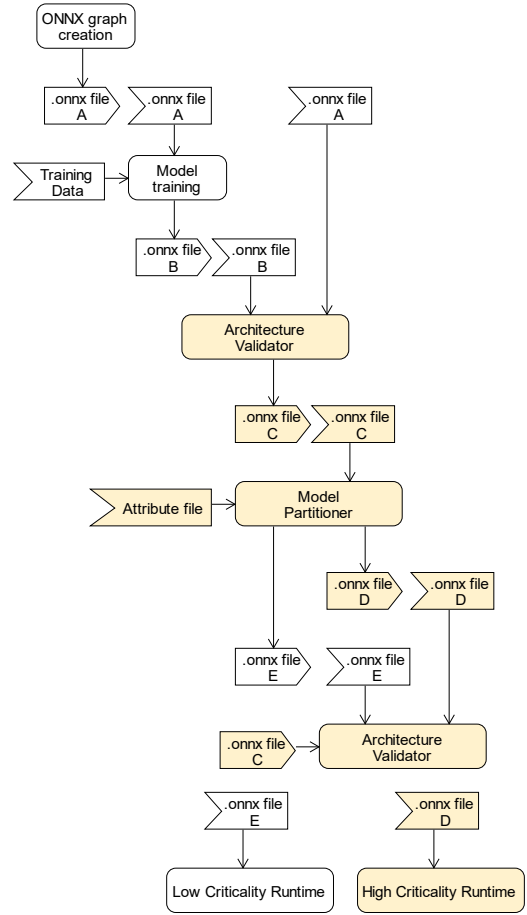

*Figure 2: Proposed AI Workflow for Mixed-Criticality AI Classifier Design and Generation. Areas shaded yellow are considered dependable.*